\documentclass{article} 
\usepackage[final]{colm2026_conference}

\usepackage{todonotes}
\usepackage{microtype}
\usepackage{hyperref}
\usepackage{url}
\usepackage{booktabs}
\usepackage{multirow}
\usepackage{makecell}
\usepackage{graphicx}
\usepackage{tabularx}
\usepackage{amsmath}
\usepackage{threeparttable}

\newcommand{\mci}[3]{\makecell[c]{#1\\[-1pt]\scriptsize[#2, #3]}}

\usepackage{lineno}

\definecolor{darkblue}{rgb}{0, 0, 0.5}
\hypersetup{colorlinks=true, citecolor=darkblue, linkcolor=darkblue, urlcolor=darkblue}

\title{Superficial Beliefs in LLM Decision-Making}

\author{Gabriel Freedman \& Francesca Toni \\
Department of Computing\\
Imperial College London\\
\texttt{\{g.freedman22,ft\}@imperial.ac.uk}
}

\begin{document}

\ifcolmsubmission
\linenumbers
\fi

\maketitle

\begin{abstract}

We ask whether large language models (LLMs) merely imitate rationales when choosing between two options, or whether their choices reflect a systematic underlying decision structure. Using synthetic binary decision settings in which models choose between profiles defined by graded attributes, we compare the attribute a model says mattered most with the attribute that best explains its choice under a behavioural model fit to prior decisions. The behavioural model predicts held-out choices well, showing that model behaviour is systematically related to the visible attributes rather than being random. However, direct self-reports and a separate score-based judge recover the behaviourally inferred driver only partially. The resulting picture reflects neither arbitrary behaviour nor fully articulated belief -- outputs are structured enough to support prediction, but explicit reasons track the recovered driver only imperfectly. This qualitative pattern persists across prompt-order and sampling perturbations, alternative behavioural models, targeted occlusion analyses, and structurally varied decision settings. We interpret this as evidence for ``superficial belief'' in LLM decision-making: models behave as if guided by probabilistic local priorities over attributes, while having only limited verbal access to the attributes that drive their decisions. Code and data can be found at: https://github.com/GIFRN/superficial\_beliefs.

\end{abstract}

\section{Introduction}

Do large language models merely imitate the language of belief, or do they sometimes deserve belief-like interpretation? Recent philosophical work has sharpened this question. From a superficialist perspective, whether a system believes something depends on stable patterns in its outward behaviour and related dispositions rather than on its underlying architecture \citep{schwitzgebel2025superficialism}. On this approach, the issue is not settled by asking what a model is made of, but by asking what sort of stable pattern its words and actions display.

This paper takes up a restricted version of that question. We do not ask whether LLMs have beliefs in general. We ask whether, in simple decision settings, their choices exhibit a local structure strong enough to support a weak form of ``superficial belief''. If models respond to similar attribute differences in similar ways, and those tendencies generalise to new cases, then there is at least a prima facie case for treating their decisions as guided by belief-like priorities rather than as arbitrary outputs. A further question is whether those priorities are also available through explicit elicitation.

This matters because LLMs not only make decisions, but can justify them. A model can state the \emph{most important} aspect of its decision, with the reasoning only loosely connected to what actually drove the choice \citep{turpin2023language,chen2025reasoningmodelsdontsay}. Our aim is to compare these two phenomena directly. We construct a benchmark of binary decisions between two options described by four graded attributes. We use the choices made by the LLMs on these decision problems to fit simple behavioural models. We then compare the models' predictions -- both of the option and the attribute driving the choice -- with those elicited directly by prompting the LLM and also using a score-based judge approach.

\begin{figure}
    \centering
    \includegraphics[width=\linewidth,clip,trim=0 70 0 70]{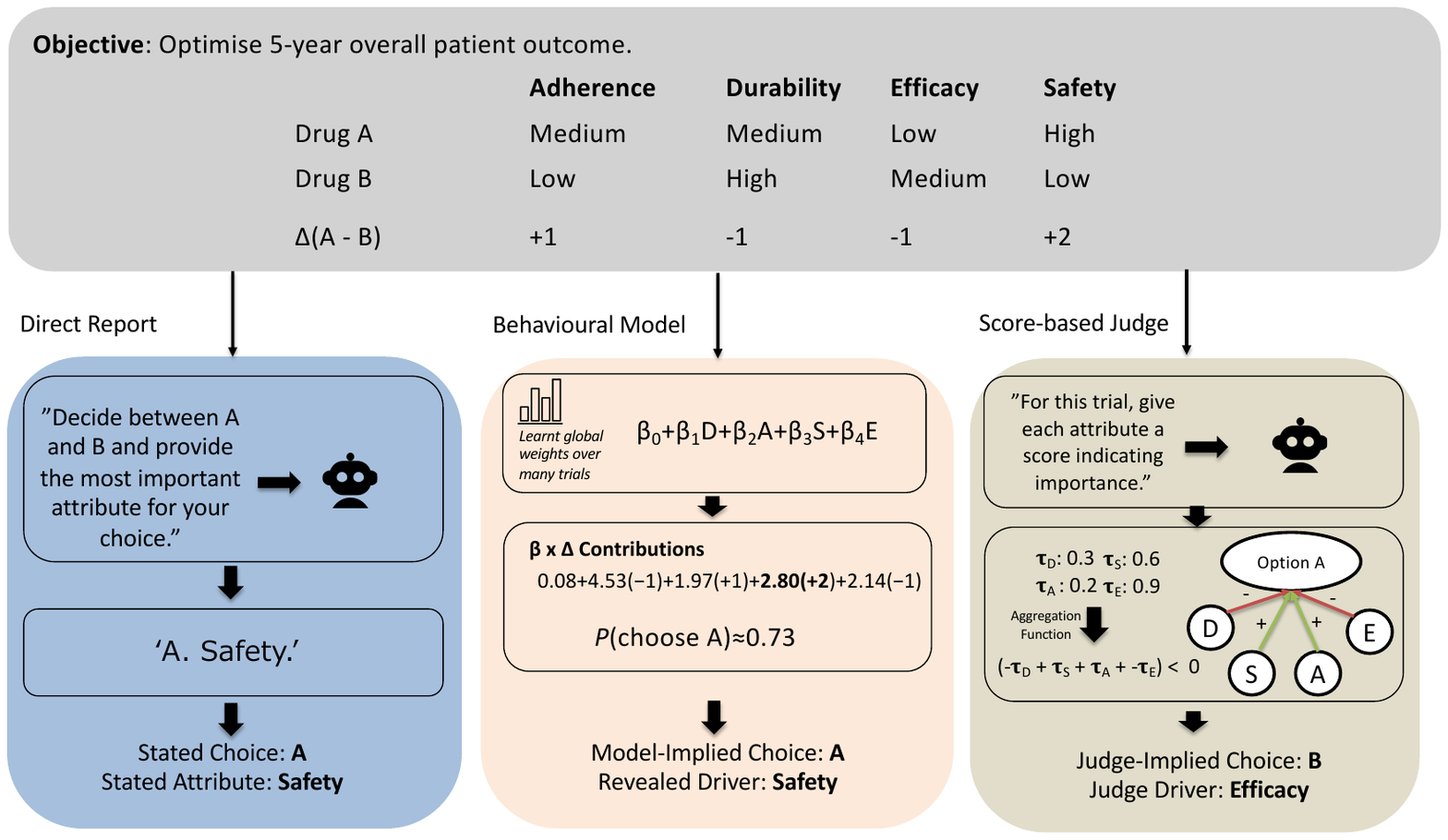}
    \caption{Illustration of a single real sample and outputs, from the \emph{Drugs} theme. The prompts are abbreviated for space, but all other values are genuine from GPT-5-mini in the non-thinking (NT) setting. (For details see Section~\ref{sec:methods}).}
    \label{fig:drug_sample}
\end{figure}

The results support a middle position. The behavioural model predicts held-out choices well across themes and model settings, indicating that the choices are systematic enough to support a revealed local structure. At the same time, agreement at the attribute level is much weaker: the attribute the model names as most important aligns with reconstructed decision drivers only partially. Control attributes are rarely selected, and the same qualitative pattern survives perturbation, intervention, and robustness checks under alternative reconstructions and structurally varied decision settings. The overall picture is therefore neither one of random behaviour nor one of fully articulated belief. Instead, we argue, it is consistent with a weak, decision-local form of superficial belief in LLM decision-making.

Concretely, we make three contributions. First, we introduce a synthetic benchmark of binary decision problems, built from a shared attribute structure and instantiated across multiple themes, including control variants. Second, we propose an operational methodology that fits simple behavioural models to LLM choices in order to reconstruct local decision priorities over attributes, and then compares those reconstructed priorities on held-out problems against two explicit elicitation methods, \emph{direct response} and \emph{score-based judge} (see Figure~\ref{fig:drug_sample}). Finally, across three main themes, two control themes, four model families, and eight total settings, together with perturbation analyses, alternative behavioural reconstructions, and a structurally different setting, we find that held-out choices are more aligned than the attributes driving those choices.


\section{Related Work}

\subsection{LLM values, preferences, and revealed decision tendencies}

A growing field of research studies whether LLMs exhibit stable value orientations or preference structures across dilemmas, surveys, and contextualised decisions. \citet{chiu2025dailydilemmas} use everyday moral quandaries to reveal which values LLMs prioritise, and \citet{rozen2025do} investigate whether LLMs display coherent value profiles, finding that apparent consistency depends strongly on elicitation strategy. Similarly, \citet{shen-etal-2025-mind} examine the misalignment between stated values and value-informed actions, while \citet{gu2025alignment} compare stated and revealed preferences directly, and likewise find substantial divergence. Others have approached the relationship between model values and belief through the lens of cognitive science \citep{murthy2026using} and utility theory \citep{mazeika2025utility}.

This literature motivates treating LLM outputs as evidence of underlying priorities, but it largely focuses on aggregate value profiles, cross-prompt stability, or broad stated-versus-revealed discrepancies. By contrast, we focus on implicitly arising behavioural dispositions, and compare them with explicit elicitations from the same models.

\subsection{Self-knowledge, belief measurement, and faithfulness of stated reasons}

\begin{figure}
    \centering
    \includegraphics[width=\linewidth,clip,trim=0 70 0 70]{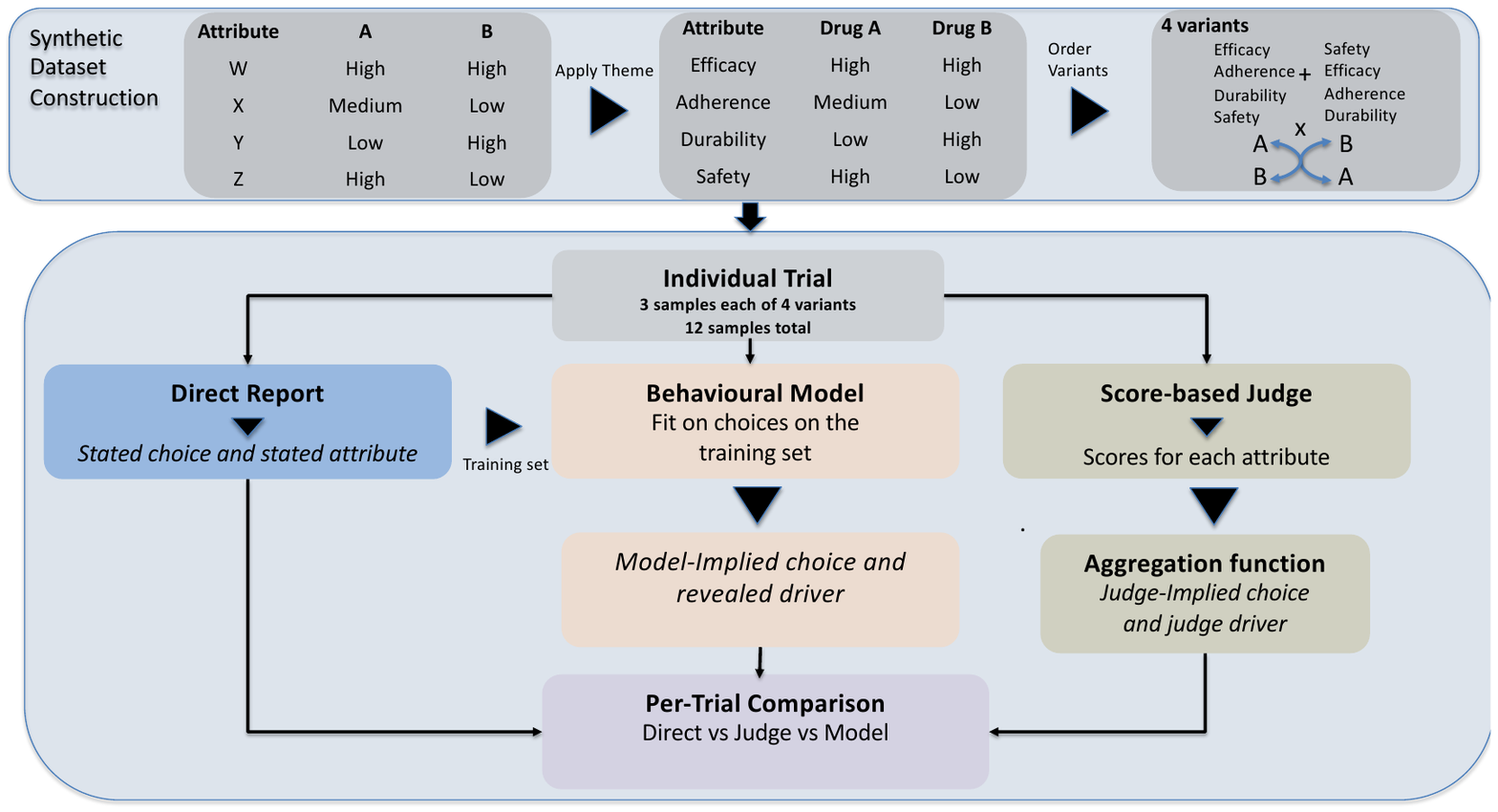}
    \caption{Overview of the entire pipeline.}
    \label{fig:pipeline_overview}
\end{figure}

Another relevant research area concerns whether models can accurately report their own epistemic states or decision processes. \citet{kadavath2022languagemodelsmostlyknow} show that LLMs can often estimate the likelihood that their answers are correct, suggesting some degree of self-knowledge at the level of uncertainty. However, recent research suggests that models do not have the ability to report their credence in a given claim in a way that satisfies basic axioms of probability \citep{freedman2025exploring}. At a more conceptual level, \citet{da76c30a25d642a6bc0a16b5b8c1904c}  argue that belief attribution should satisfy conditions such as accuracy, coherence, uniformity and use, rather than resting on a single verbal report. This perspective is directly relevant to our setting, where we not only ask whether a model can produce a plausible explanation, but whether that explanation is connected to the structure of its behaviour.

\section{Methods}
\label{sec:methods}

\subsection{Benchmark design}

The benchmark begins with 400 training and 100 test source problems, each defined by a pair of four-attribute profiles. These source problems are then rendered in five themes, and each theme-conditioned problem is expanded into four prompt variants. Each theme is associated with an objective included in the prompts.

\subsubsection{Base configurations}
\label{sec:congifs}

In the base configurations, each of
four attributes (W, X, Y, Z, before theme labels are applied) takes one of three ordered levels: \emph{low}, \emph{medium}, or \emph{high}. Each source problem is built from two different four-attribute profiles, \(P\) and \(Q\), sampled over the three levels. We keep only pairs where each profile is better on at least one attribute, so every problem contains a real trade-off rather than a dominant option. For each theme, we generate 400 such source problems for training and 100 for testing
.

We use 
a restricted ordering scheme, comprising four fixed global orders, specifying the order the attributes are displayed in 
each prompt, as follows:
\[
[W,X,Y,Z],\quad [X,W,Z,Y],\quad [Z,Y,W,X],\quad [Y,Z,X,W].
\]
Each source problem inherits two of these orders -- meaning the attributes for both profiles are presented in that order. For each inherited order, we create two prompt versions: one with profile \(P\) shown as option \(A\) and \(Q\) as option \(B\) (whether a profile is `A' or `B' dictates whether it is shown first or second in the prompt, see Appendix \ref{app:prompts} for full prompts) and one with those labels reversed. Thus, each profile appears equally often as both options. For example, if one inherited order is \([W,X,Y,Z]\) and the other \([X,W,Z,Y]\), both displayed options are shown as \([W,X,Y,Z]\) once with \(P\) as \(A\) and \(Q\) as \(B\), and once 
with \(Q\) as \(A\) and \(P\) as \(B\). Repeating this for \([X,W,Z,Y]\) gives four rendered prompt rows per source problem.

\subsubsection{Theme application}

We use three 
themes, i.e., drugs, policy, and software, and two control versions of the drugs theme. Across the three main themes, the objective provided to the LLM within prompts is to choose the option that best improves, respectively, a five-year patient outcome, a five-year community outcome, or a five-year production-engineering outcome for a small team.

The three main themes use the same underlying four attribute structure (see Section~\ref{sec:congifs}), but with different attribute names (Table~\ref{tab:theme-slots}). We deliberately choose objectives and attributes that are ambiguous and open to interpretation, to avoid unintentionally implying a single `correct' answer. The two control themes keep the drugs framing and the first three drug attributes unchanged, but replace \emph{Durability} with one of two attributes which are chosen to be irrelevant to the objective: either \emph{Packaging Symmetry} or \emph{Label Border Thickness}. Each control remains a four-attribute task, with the irrelevant attribute replacing \emph{Durability} in the same table position and using its corresponding \emph{low}/\emph{medium}/\emph{high} levels, rather than being added as a fifth attribute.

\begin{table}[t]
\centering
\small
\setlength{\tabcolsep}{3pt}
\renewcommand{\arraystretch}{1.05}
\begin{tabularx}{\columnwidth}{@{}>{\raggedright\arraybackslash}X>{\raggedright\arraybackslash}X>{\raggedright\arraybackslash}X@{}}
\toprule
Drugs & Policy & Software \\
\midrule
Efficacy & Effectiveness & Capability \\
Adherence & Compliance & Adoption Ease \\
Safety & Safety & Reliability \\
Durability & Implementation Ease & Maintainability \\
\bottomrule
\end{tabularx}
\caption{Theme-specific attribute names. Reading down each column gives the four attributes used in that theme. In the two control themes \emph{Durability} is replaced with \emph{Packaging Symmetry} or \emph{Label Border Thickness} within the Drug theme.}
\label{tab:theme-slots}
\end{table}

Each theme-conditioned decision problem is sampled three times. As a result, for each theme and each model condition (as detailed below), a full training run contains 1600 prompt rows ($400 \times 4$ variants, see Section~\ref{sec:congifs})  and 4800 responses (\(1600 \times 3\) samples), while a full test run contains 400 prompt rows and \(400 \times 3 = 1200\) responses.

\subsection{Model families and prompts}

We study four model families, each in two settings, giving eight conditions in total: GPT-5-mini, GPT-5-nano \citep{singh2025openaigpt5card}, Qwen3-14B \citep{yang2025qwen3technicalreport}, and Ministral-3-14B \citep{liu2026ministral3}, each in a non-thinking (NT) and a thinking (T) setting. For GPT-5-mini and GPT-5-nano, this distinction is realised through the reasoning effort parameter. Likewise, for Qwen3-14B, it is realised by disabling versus enabling thinking mode. For Ministral-3-14B, it is realised by switching checkpoints (the 
NT setting uses an Instruct checkpoint, whereas the 
T setting uses a Reasoning checkpoint). The labels NT and T therefore describe the practical behaviour of the settings rather than an identical mechanism.

For the first explicit elicitation strategy, the model is instructed to optimise the theme objective, choose exactly one option, and then name the single most important attribute for that choice. The required output is a parseable and structured line, for example \texttt{`A. Safety.'} or \texttt{`B. Capability.'}. We refer to this as the \emph{direct response}.  For the second elicitation strategy, we query the same model family and setting again in a fresh context, but instead of asking for a choice, we ask for one decisiveness score in \([0,1]\) for each of the four attributes. Full prompts can be seen in Appendix \ref{app:prompts}. The required output is four lines of the form \texttt{<Attribute>=<[0,1]>}, e.g. \texttt{Safety=0.7}. We refer to this as the \emph{score-based judge}. These scores are then directionalised using the visible sign of the corresponding attribute difference, as described in Section \ref{sec:ArgLLM}.


\subsection{Behavioural model}

For each theme, model family, and setting, we fit a separate binomial logistic regression to the training split. Let \(F=\{W,X,Y,Z\}\) denote the four attribute positions. For each rendered training row \(t\), let \(y_t\in\{0,1,2,3\}\) be the number of times the model chose \(A\) across the three repeated samples. We model
\[
y_t \sim \mathrm{Binomial}(3,p_t),
\]
where \(p_t\) is the probability of choosing \(A\) on prompt row \(t\).

The predictors are the four visible attribute differences. Each attribute is shown at one of three ordered levels (\emph{low}, \emph{medium}, or \emph{high}) which we encode as \(-1\), \(0\), and \(1\). For row \(t\) and attribute \(f\), let \(x^A_{tf}\) and \(x^B_{tf}\) denote the encoded levels shown for options \(A\) and \(B\). We then define the visible attribute difference as
\[
d_{tf} = x^A_{tf} - x^B_{tf},
\]
so \(d_{tf} \in \{-2,-1,0,1,2\}\). Positive values mean \(A\) is better than \(B\) on attribute \(f\), negative values mean the reverse, and zero means that the two attributes are equal.

Concretely, we write
\[
\operatorname{logit}(p_t)
=
\beta_0
+
\sum_{f\in F}\beta_f\, d_{tf},
\]
where \(p_t\) is the probability of choosing \(A\) on row \(t\). The sum captures how strongly each attribute difference affects the choice.

On the test split, the fitted model gives both a predicted choice and one signed contribution per attribute. For attribute \(f\) on row \(t\), we define
\[
C_{tf}
=
\beta_f\, d_{tf}.
\]
The quantity \(C_{tf}\) is the contribution of attribute \(f\) to the model's preference for \(A\) over \(B\): positive values favour \(A\), negative values favour \(B\). For the main four-attribute benchmark and alternative-surrogate analyses, we define the revealed driver relative to the behavioural model's predicted choice on that row. Let \(\hat y_t\in\{A,B\}\) denote this predicted choice. We 
set
\[
f_t^\star
=
\begin{cases}
\arg\max_{f\in F} C_{tf}, & \text{if } \hat y_t=A,\\[4pt]
\arg\min_{f\in F} C_{tf}, & \text{if } \hat y_t=B.
\end{cases}
\]
This is the attribute that most strongly supports the side predicted by the behavioural model. We treat this behavioural model as an operational reconstruction of local decision structure, not as direct access to ground truth. For the margin, mismatch-case intervention, and hospital analyses, we use an observed-choice-conditioned driver \(f_t^{\star,\mathrm{obs}}\), obtained by applying the same rule to the observed direct-response choice \(y_t^{\mathrm{obs}}\in\{A,B\}\). For the margin-conditioned analysis below, we orient contributions toward the observed choice by setting \(R_{tf}=C_{tf}\) when \(y_t^{\mathrm{obs}}=A\) and \(R_{tf}=-C_{tf}\) when \(y_t^{\mathrm{obs}}=B\). We define the behavioural margin as \(m_t=R_{t,(1)}-R_{t,(2)}\), the gap between the largest and second-largest chosen-side contributions.

\subsection{ArgLLM-style score-based judge}
\label{sec:ArgLLM}

As an alternative to the direct response baseline, we follow the ArgLLM \citep{Freedman_Dejl_Gorur_Yin_Rago_Toni_2025} idea of evaluating the components of a decision separately, but apply it here to an implicitly constructed quantitative bipolar argumentation framework rather than to an explicitly verbalised graph. For each row \(t\) and each attribute \(f\in F\), the score-based judge returns a raw decisiveness score \(\tau_{tf}\in[0,1]\). Intuitively, \(\tau_{tf}\) is the strength assigned to the attribute difference on \(f\), before taking into account which option that difference favours.

We then make each score either positive or negative using the visible sign of the underlying attribute difference. Let
\[
s_{tf}
=
\begin{cases}
\phantom{-}1, & \text{if attribute } f \text{ favours } A,\\
-1, & \text{if attribute } f \text{ favours } B,\\
\phantom{-}0, & \text{if the two options are tied on } f.
\end{cases}
\]
The signed contribution of attribute \(f\) is
\[
\tilde{\tau}_{tf} = s_{tf}\tau_{tf}.
\]
Positive values therefore support \(A\), negative support \(B\), and zero means neutral support.

For the main results, we use a simple signed-sum aggregation function (formally known in the literature as an \emph{argumentation semantics}~\citep{Freedman_Dejl_Gorur_Yin_Rago_Toni_2025}):
\[
S_t = \sum_{f\in F}\tilde{\tau}_{tf}.
\]
The evaluator predicts \(A\) if \(S_t \ge 0\) and \(B\) otherwise; exact ties are broken in favour of \(A\). Because the benchmark is balanced over the two \(A/B\) labellings, this deterministic tie-break does not create a directional bias. For attribute-level evaluation, however, we do not condition on the judge-implied choice. Instead, we define the score-based judge's attribute output as the signed contribution with the largest absolute value:
\[
g_t^\star = \arg\max_{f\in F}\left|\tilde{\tau}_{tf}\right|.
\]

\subsection{Targeted occlusion validation}

To explore alternate approaches, we report a focused occlusion study in one illustrative setting, GPT-5-mini non-thinking on the drugs theme. Starting from 400 test problems, we create one baseline version and, for each of the four attributes, one \emph{equalise} version and one \emph{drop} version, giving 3600 rows in total.

In the \emph{drop} intervention, the visible evidence for one attribute is removed, and for \emph{equalise} it is set to \emph{medium} for both of the choices, with the rest of the attribute levels left unchanged. We use this suite to test whether attributes ranked higher by the behavioural model produce larger changes in the model's choice and attribute selection when perturbed.

\begin{table*}[t]
\centering
\footnotesize
\setlength{\tabcolsep}{3.6pt}
\renewcommand{\arraystretch}{1.08}
\begin{threeparttable}

\begin{tabular*}{\textwidth}{@{\extracolsep{\fill}}>{\raggedright\arraybackslash}p{3.1cm}cccccccc@{}}
\toprule
& \multicolumn{4}{c}{Direct response} & \multicolumn{4}{c@{}}{Score-based judge} \\
\cmidrule(r){2-5}\cmidrule(l){6-9}
& \multicolumn{2}{c}{Choice} & \multicolumn{2}{c}{Attribute} & \multicolumn{2}{c}{Choice} & \multicolumn{2}{c@{}}{Attribute} \\
\cmidrule(r){2-3}\cmidrule(lr){4-5}\cmidrule(lr){6-7}\cmidrule(l){8-9}
Setting & NT & T & NT & T & NT & T & NT & T \\
\midrule

\multicolumn{9}{@{}l}{\textit{Drugs}}\\
\hspace{0.8em}GPT-5-mini        & 0.878 & 0.816 & 0.665 & 0.722 & 0.871 & 0.876 & 0.787 & 0.766 \\
\hspace{0.8em}GPT-5-nano        & 0.800 & 0.797 & 0.511 & 0.603 & 0.694 & 0.838 & 0.478 & 0.587 \\
\hspace{0.8em}Qwen3-14B         & 0.756 & 0.775 & 0.512 & 0.633 & 0.758 & 0.753 & 0.515 & 0.627 \\
\hspace{0.8em}Ministral-3-14B   & 0.759 & 0.652 & 0.528 & 0.420 & 0.730 & 0.796 & 0.638 & 0.664 \\
\addlinespace[2pt]
\textit{Drugs aggregate}
  & \multicolumn{2}{c}{\mci{0.779}{0.762}{0.795}}
  & \multicolumn{2}{c}{\mci{0.574}{0.551}{0.599}}
  & \multicolumn{2}{c}{\mci{0.789}{0.767}{0.813}}
  & \multicolumn{2}{c@{}}{\mci{0.633}{0.608}{0.655}} \\

\addlinespace[4pt]
\multicolumn{9}{@{}l}{\textit{Policy}}\\
\hspace{0.8em}GPT-5-mini        & 0.887 & 0.890 & 0.804 & 0.852 & 0.802 & 0.811 & 0.810 & 0.825 \\
\hspace{0.8em}GPT-5-nano        & 0.814 & 0.858 & 0.554 & 0.750 & 0.677 & 0.764 & 0.468 & 0.707 \\
\hspace{0.8em}Qwen3-14B         & 0.887 & 0.868 & 0.703 & 0.789 & 0.727 & 0.804 & 0.460 & 0.732 \\
\hspace{0.8em}Ministral-3-14B   & 0.792 & 0.662 & 0.663 & 0.458 & 0.766 & 0.790 & 0.664 & 0.669 \\
\addlinespace[2pt]
\textit{Policy aggregate}
  & \multicolumn{2}{c}{\mci{0.832}{0.817}{0.846}}
  & \multicolumn{2}{c}{\mci{0.697}{0.673}{0.719}}
  & \multicolumn{2}{c}{\mci{0.768}{0.744}{0.793}}
  & \multicolumn{2}{c@{}}{\mci{0.667}{0.644}{0.691}} \\

\addlinespace[4pt]
\multicolumn{9}{@{}l}{\textit{Software}}\\
\hspace{0.8em}GPT-5-mini        & 0.859 & 0.859 & 0.617 & 0.772 & 0.853 & 0.626 & 0.608 & 0.583 \\
\hspace{0.8em}GPT-5-nano        & 0.865 & 0.752 & 0.368 & 0.552 & 0.522 & 0.849 & 0.317 & 0.623 \\
\hspace{0.8em}Qwen3-14B         & 0.858 & 0.803 & 0.594 & 0.647 & 0.783 & 0.814 & 0.397 & 0.671 \\
\hspace{0.8em}Ministral-3-14B   & 0.761 & 0.646 & 0.497 & 0.429 & 0.721 & 0.783 & 0.583 & 0.548 \\
\addlinespace[2pt]
\textit{Software aggregate}
  & \multicolumn{2}{c}{\mci{0.800}{0.786}{0.815}}
  & \multicolumn{2}{c}{\mci{0.559}{0.536}{0.583}}
  & \multicolumn{2}{c}{\mci{0.744}{0.722}{0.766}}
  & \multicolumn{2}{c@{}}{\mci{0.541}{0.518}{0.563}} \\

\midrule
\textbf{All themes aggregate}
  & \multicolumn{2}{c}{\mci{\textbf{0.804}}{0.794}{0.812}}
  & \multicolumn{2}{c}{\mci{\textbf{0.610}}{0.597}{0.624}}
  & \multicolumn{2}{c}{\mci{\textbf{0.767}}{0.754}{0.781}}
  & \multicolumn{2}{c@{}}{\mci{\textbf{0.613}}{0.599}{0.628}} \\
\bottomrule
\end{tabular*}

\vspace{8pt}

\begin{tabular*}{\textwidth}{@{\extracolsep{\fill}}lccc@{}}
\toprule
\multicolumn{4}{@{}l}{\textbf{Control themes}}\\
\addlinespace[2pt]
 & DR = control & BM = control & SBJ = control \\
\midrule
Packaging symmetry
  & \mci{0.003}{0.002}{0.004}
  & \mci{0.016}{0.010}{0.023}
  & \mci{0.013}{0.009}{0.019} \\

Label border thickness
  & \mci{0.002}{0.001}{0.003}
  & \mci{0.011}{0.006}{0.018}
  & \mci{0.011}{0.007}{0.016} \\
\bottomrule
\end{tabular*}

\caption{
\textbf{Benchmark summary, model-by-setting breakdown, and control-theme checks.} In the main block, \emph{Direct response / Choice} is agreement between the behavioural model's predicted choice and the model's observed choice. \emph{Direct response / Attribute} is agreement between the behavioural model's revealed driver \(f_t^\star\), defined relative to the behavioural model's predicted choice, and the attribute named by the model. \emph{Score-based judge / Choice} is agreement between the score-based judge's reconstructed choice and the same behavioural-model choice target. \emph{Score-based judge / Attribute} is agreement between the score-based judge's largest-magnitude signed contribution \(g_t^\star\) and the same behavioural-model revealed driver \(f_t^\star\). Each theme block reports the four model families under non-thinking (NT) and thinking (T) settings, followed by that theme's aggregate across all eight model-setting conditions. The final row is the aggregate over all three main themes. DR = direct response, BM = behavioural model, and SBJ = score-based judge. Control theme rows and theme and overall aggregates report 95\% CIs, model-by-setting rows report point estimates only.}
\label{tab:main_results}

\end{threeparttable}
\end{table*}

\section{Results}

\begin{figure}
    \centering
\includegraphics[width=\linewidth,clip,trim=0 90 0 95]{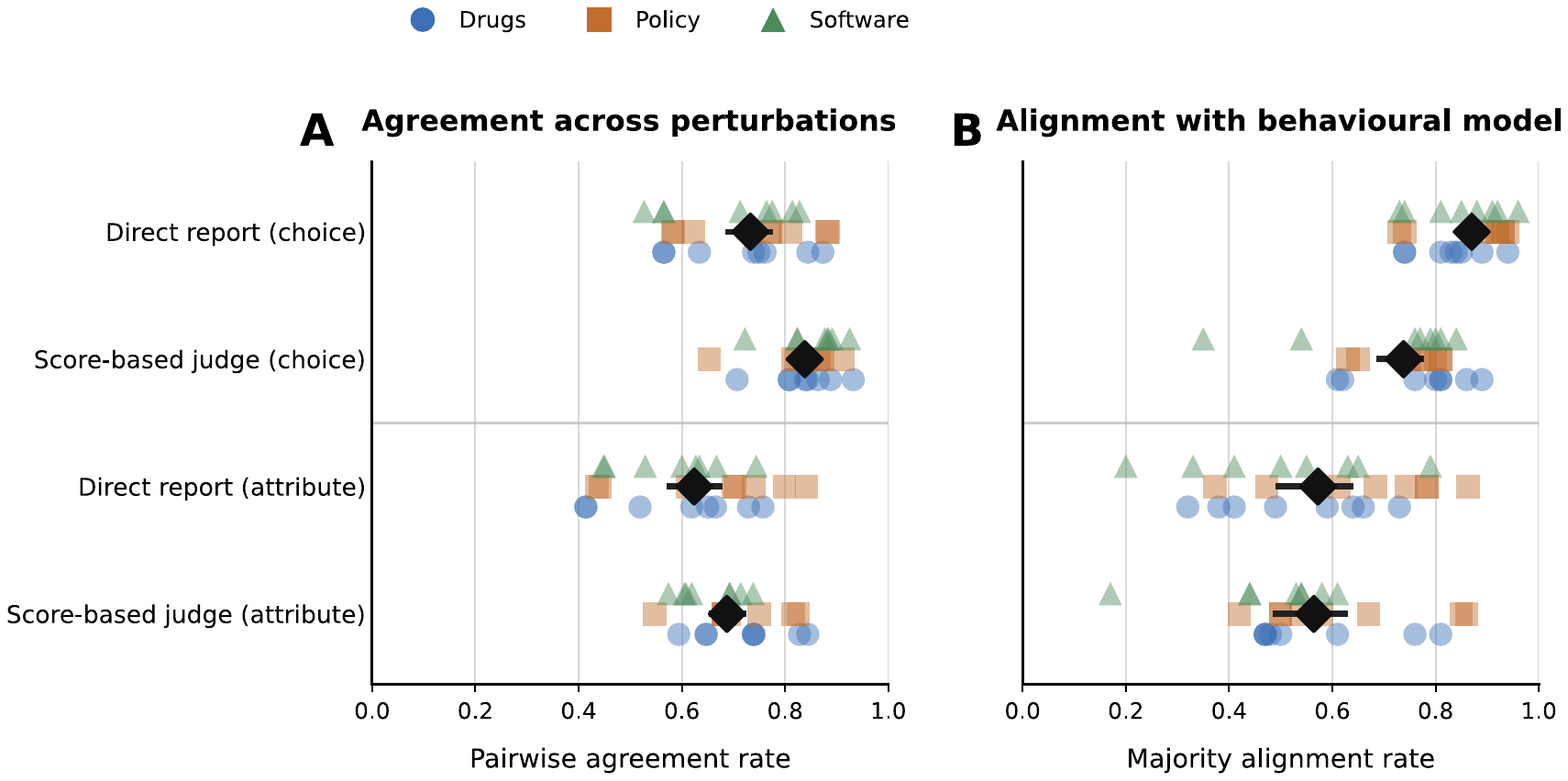}
    \caption{Agreement across perturbations (left) and alignment with the behavioural model (right) for direct responses and score-based judge outputs at the choice and attribute levels. Faded points show individual conditions and diamonds show pooled means with 95\% CIs.}
    \label{fig:reproducibility_alignment}
\end{figure}

We report three kinds of evidence. Table~\ref{tab:main_results} gives per-draw agreement with behavioural-model targets. Figure~\ref{fig:reproducibility_alignment} then asks how concentrated responses remain under prompt-order and sampling perturbations, and how often a base configuration's dominant response matches the same targets. Figure~\ref{fig:occlusion_figure} provides an illustrative intervention analysis for one model-theme setting. Appendix analyses extend this with alternative behavioural reconstructions, additional intervention settings, and a structurally different decision setting.

\subsection{Choice prediction and attribute recovery}

Table~\ref{tab:main_results} shows that the behavioural model predicts held-out choices well across all three main themes. Aggregating across themes and model settings, it matches the model's own choice \(80.4\%\) of the time \([79.4,81.2]\). The score-based judge, evaluated against the same behavioural model choice target, reaches \(76.7\%\) \([75.4,78.1]\). Direct response with the behavioural model is strongest in policy (\(83.2\%\)) and remains high in drugs (\(77.9\%\)) and software (\(80.0\%\)). Judge-choice agreement is slightly higher than direct-choice agreement in drugs and lower in policy and software. Table~\ref{tab:simple_choice_baselines} in the Appendix shows that this prediction capability cannot be recreated by simple heuristics alone, as an equal-weight additive rule and a count-better rule are 11\% and 17\% less accurate on held-out choice prediction on aggregate.

At the attribute level, agreement is weaker and more mixed. The attribute named by the model matches the behavioural model's revealed driver only partially, with an overall rate of \(61.0\%\) \([59.7,62.4]\). The corresponding score-based judge rate, evaluated against the same revealed driver target, is very similar overall at \(61.3\%\) \([59.9,62.8]\), though the relative ordering varies by theme. In drugs, the score-based judge is somewhat higher than the direct report (\(63.3\%\) versus \(57.4\%\)), in policy it is slightly lower (\(66.7\%\) versus \(69.7\%\)), and in software the two are again close (\(54.1\%\) versus \(55.9\%\)). The main result is therefore not that one explicit method clearly outperforms the other, but that both provide only partial access to the decision structure recovered from behaviour.

It is worth noting that a strict top-1 comparison can overstate disagreement when the two largest behavioural contributions are similar. In the lowest-margin quartile, direct-report disagreement with the observed-choice-conditioned behavioural driver is \(42.0\%\), and score-based-judge disagreement is \(55.3\%\). In the highest-margin quartile, these rates fall to \(11.0\%\) and \(18.8\%\), respectively. Thus, some top-1 mismatches reflect close multi-attribute trade-offs, while disagreement remains in a non-trivial minority of cases with a clearly separated top driver.

The control themes show that these findings are not simply a matter of models arbitrarily naming attributes. In the two control versions of the drugs theme, the irrelevant attribute is selected only rarely: direct reports choose it in \(0.2\%\) to \(0.3\%\) of cases, the behavioural model recovers it in about \(1.1\%\) to \(1.6\%\), and the score-based judge in about \(1.1\%\) to \(1.3\%\). These low rates support the interpretation that the benchmark is exhibiting semantically meaningful structured preferences, rather than choosing based on entirely random artefacts. Furthermore, Appendix~\ref{app:behavioural-robustness} shows that the headline results are stable under two alternative behavioural models. The same qualitative pattern appears in the six-attribute hospital setting (Appendix~\ref{app:hospital-setting}): held-out choice agreement is \(84.7\%\) for GPT-5-mini NT and \(89.0\%\) for Qwen3-14B NT, while direct attribute agreement is \(64.4\%\) and \(56.0\%\), respectively.

\subsection{Perturbation robustness and intervention evidence}

Figure~\ref{fig:reproducibility_alignment} asks a different question. Instead of looking at individual outputs, it summarises families of twelve related realisations of the same underlying problem: four prompt variants and three samples per variant. Panel A shows that the response streams are not perfectly invariant, but they are far from random. At the pooled level, the score-based judge is more reproducible than the direct response stream for both choices and attributes. Within-family pairwise agreement is \(0.838\) \([0.808, 0.863]\) for score-based judge choices, compared with \(0.733\) \([0.685, 0.777]\) for direct choices, and \(0.687\) \([0.653, 0.725]\) for score-based judge attributes, compared with \(0.624\) \([0.570, 0.678]\) for directly stated attributes. 

Panel B shows that reproducibility and recovery of the behavioural model targets come apart. Direct choices align more strongly with the behavioural model choice target than score-based judge choices, with majority correct rates of \(0.870\) \([0.838, 0.896]\) and \(0.738\) \([0.686, 0.778]\), respectively. At the attribute level, however, the two explicit methods are almost indistinguishable: direct reports recover the behavioural model driver at a rate of \(0.572\) \([0.490, 0.640]\), while the score-based judge reaches \(0.564\) \([0.484, 0.631]\). This suggests not that the score-based judge is uniformly worse, but that greater reproducibility does not by itself imply closer correspondence to the decision structure. Together, these results support the position that outputs are concentrated enough to reveal a meaningful local tendency, but that tendency is not fully stable across perturbations and is only partially accessible to explicit report. Appendix perturbation summaries in the structurally different setting show the same broad dissociation. Score-based outputs can be more reproducible than direct reports while remaining less behaviourally aligned, especially at the attribute level.

\begin{figure}
    \centering
    \includegraphics[width=1\linewidth]{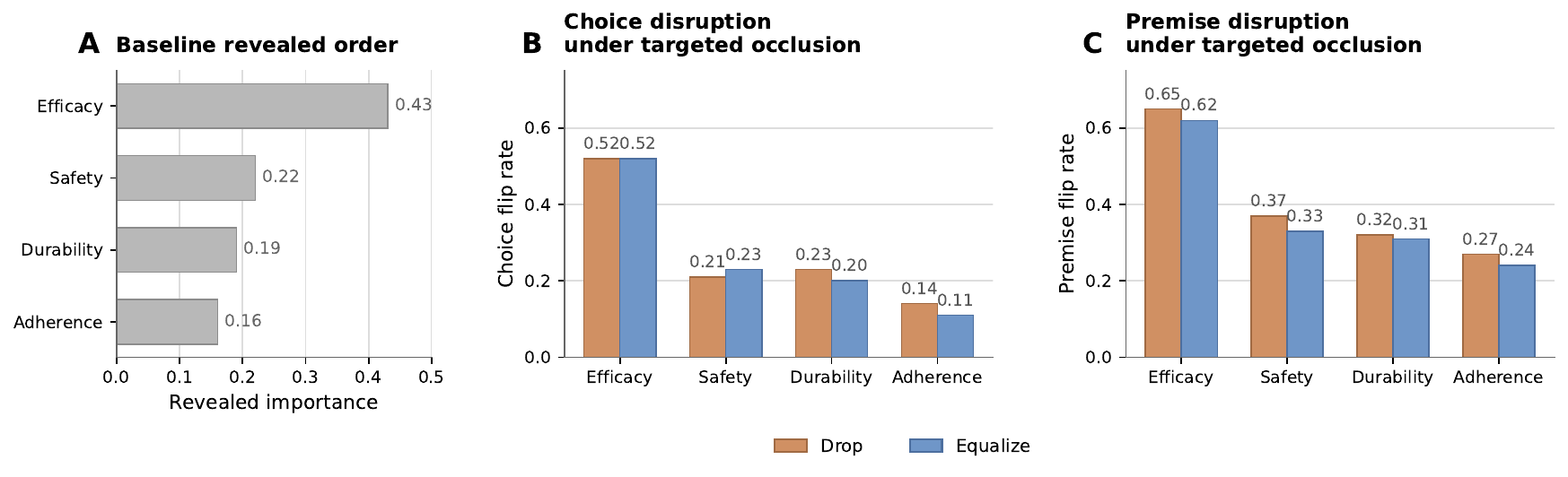}
   \caption{Panel A shows the baseline attribute order inferred from matched baseline choices in the drugs theme ($E \textgreater S \textgreater D \textgreater A$) for GPT‑5‑mini NT. Panels B and C show, respectively, choice-flip rates and stated-attribute flip rates when evidence for each attribute is either removed (\emph{drop}) or neutralised (\emph{equalise}) while the rest of the matched item family is held fixed. More important baseline attributes produce larger disruptions to both choices and stated attributes.}
    \label{fig:occlusion_figure}
\end{figure}

Figure~\ref{fig:occlusion_figure} provides a more direct test of whether the recovered ranking matters for behaviour in one illustrative setting, focusing on GPT-5-mini NT on the drugs theme. Panel A shows the baseline ranking inferred from the behavioural model fit on the training set. Panels B and C then intervene on each attribute in turn, either by removing the visible evidence for that attribute (\emph{drop}) or by neutralising it (\emph{equalise}). Attributes ranked higher by the behavioural model produce larger disruptions in both choices and stated attributes compared to those lower ranked. This does not by itself establish a benchmark-wide causal claim, but it does provide targeted corroboration that the behavioural model is recovering structure that is relevant to what the model actually does when deciding. Appendix intervention analyses broaden this beyond the illustrative case -- two further four-attribute settings show the same qualitative top-rank sensitivity pattern (Appendix \ref{app:additional-occlusion}), and in the structurally different hospital setting \emph{equalise} occlusions again shift stated key attributes more readily than choices (Table~\ref{tab:hospital-occlusion}).

Furthermore, across the \(314\) held-out cases from the three four-attribute occlusion settings in which the directly stated driver and the observed-choice-conditioned behavioural driver disagree, intervening on the behavioural driver changes the model's choice more often. Under \emph{drop}, removing the behavioural driver changes the choice in \(186/314\) cases (\(59.2\%\)), compared with \(113/314\) (\(36.0\%\)) for the stated driver. Under \emph{equalise}, the corresponding rates are \(187/314\) (\(59.6\%\)) and \(96/314\) (\(30.6\%\)). Thus, in mismatch cases the behavioural reconstruction more often identifies the higher-impact attribute under these interventions.

\section{Discussion and Future Work}

Our results invite only a cautious claim about superficial beliefs. A first lesson is that, in the tasks we identified, LLM decisions are neither arbitrary nor fully transparent. Held-out choice prediction, low control attribute selection rates, perturbation evidence, intervention analyses, and robustness checks in a structurally different decision setting all indicate a meaningful local decision structure. But that structure is better described as probabilistic and local than as a fixed implicit preference: direct reports recover the revealed driver only partially, and exact agreement across prompt variants and repeated samples is uncommon.

The margin analysis shows that disagreement is lower when the leading contribution is more clearly separated, while the mismatch-case interventions show that the distinction between stated and behavioural drivers can still matter for the model's choices.

A second lesson is that reproducibility and recovery come apart. The (argumentative) score-based judge is more reproducible than the direct LLM responses. Its choice agreement with the behavioural model is lower overall, although the ordering varies by theme, while attribute agreement is broadly similar. This should not be seen as detrimental to the validity of argumentation-based approaches. In the original ArgLLM framework the judge is based on, the point of argumentation is to make decisions more explainable and contestable through structured intermediate representations and formal aggregation, and newer work extends this idea to more general decision support and global contestability \citep{Freedman_Dejl_Gorur_Yin_Rago_Toni_2025,dejl2026argumentationexplainablegloballycontestable}. In that light, our score-based judge may be useful less as a privileged readout of latent priorities than as a stable and contestable scaffold for decision-making.

Our findings therefore support only a weak, decision-local form of superficial belief. On a superficialist view, belief attribution depends on stable patterns in outward behaviour rather than hidden internal architecture \citep{schwitzgebel2025superficialism}. Our results fit that picture in a limited way: within synthetic binary decision settings, models behave as if guided by local priorities over visible attributes, and the same qualitative dissociation persists in a structurally different decision setting. But the evidence still falls short of stronger claims. The priorities are not perfectly invariant, they are only partly recoverable in explicit elicitation methods, and we do not show that they connect to broader world-model commitments or generalise beyond a small class of synthetic pairwise tasks. In this sense, the results are closer to a weak behavioural disposition than to a fully articulated belief state, and they remain short of stronger standards for belief attribution \citep{da76c30a25d642a6bc0a16b5b8c1904c}.

Some limitations remain concerning the scope of the study, pointing to future work. Specifically, although the main pattern survives (i) alternative behavioural reconstructions, (ii) additional intervention settings, and (iii) a structurally different decision setting, the evidence is still synthetic and centred on binary profile comparison tasks. Broader generalisation to richer and less stylised decision settings therefore remains open.

\section*{Acknowledgements}
Freedman was funded by UKRI through the CDT in Safe and Trusted Artificial Intelligence (Grant No. EP/S023356/1). Toni was funded by ERC under the EU's Horizon 2020 research and innovation programme (grant agreement No. 101020934, ADIX), and by EPSRC (grant UKRI3928, NeSyDebates).

\bibliography{colm2026_conference}
\bibliographystyle{colm2026_conference}

\appendix
\section{Supplementary Analyses}

\subsection{Appendix roadmap}
\label{app:appendix-roadmap}

This appendix broadens the main text evidence in three ways. Section~\ref{app:behavioural-robustness} tests robustness to alternative behavioural surrogates fit on the same training choices. Section~\ref{app:hospital-setting} evaluates a structurally different six-attribute decision setting under the same core protocol as the main benchmark. Section~\ref{app:additional-occlusion} reports two further targeted-occlusion validations beyond the illustrative main text case, and Section~\ref{app:stats-impl} collects statistical and implementation details. Section~\ref{app:prompts} provides the full prompts used for experiments.

\subsection{Robustness to alternative behavioural models}
\label{app:behavioural-robustness}

\subsubsection{Calibration against simple non-learned choice baselines}

To calibrate the main behavioural model against simple non-learned alternatives, Table~\ref{tab:simple_choice_baselines} compares it with two heuristic choice rules on the same held-out evaluation used for Table~\ref{tab:main_results}. The equal-weight additive rule predicts \(A\) when the sum of the signed visible attribute differences is non-negative and \(B\) otherwise. The count-better rule predicts \(A\) when at least as many attributes favour \(A\) as favour \(B\), and \(B\) otherwise.

\begin{table}[ht]
\centering
\small
\setlength{\tabcolsep}{4.5pt}
\renewcommand{\arraystretch}{1.08}
\begin{tabular}{@{}lccc@{}}
\toprule
Scope & Equal-weight additive & Count-better & Behavioural model \\
\midrule
Drugs & 0.703 & 0.624 & 0.779 \\
Policy & 0.686 & 0.638 & 0.832 \\
Software & 0.687 & 0.639 & 0.800 \\
\midrule
Aggregate & 0.692 & 0.634 & 0.804 \\
\bottomrule
\end{tabular}
\caption{Calibration against simple held-out choice baselines on the substantive themes. Entries are per-response choice agreement, that is, agreement between each rule's predicted choice and the model's observed choice under the same evaluation convention as Table~\ref{tab:main_results}.}
\label{tab:simple_choice_baselines}
\end{table}

\subsubsection{Model definitions and metric conventions}

Let \(\Delta_j \in \{-2,-1,0,1,2\}\) denote the signed advantage of option \(A\) over option \(B\) on attribute \(j\), and let \(\sigma(\cdot)\) be the logistic link. The main model, \(M0\), is the paper's linear additive surrogate,
\[
\Pr(A)=\sigma\!\left(\alpha+\sum_j \beta_j \Delta_j\right).
\]
\(M1\) relaxes linearity while remaining additive by assigning separate coefficients to one-step and two-step moves on each attribute,
\[
\Pr(A)=\sigma\!\left(\alpha+\sum_j \bigl(\beta_{j,1} z_{j,1}+\beta_{j,2} z_{j,2}\bigr)\right),
\]
where \(z_{j,1}=\mathbf{1}\{\Delta_j=1\}-\mathbf{1}\{\Delta_j=-1\}\) and \(z_{j,2}=\mathbf{1}\{\Delta_j=2\}-\mathbf{1}\{\Delta_j=-2\}\). \(M2\) is a smoothed exact cell lookup model over the canonical four attribute difference cell \(c=(\Delta_W,\Delta_X,\Delta_Y,\Delta_Z)\), with
\[
p_{M2}(A\mid c)=\frac{k_c+\lambda\,p_0(c)}{n_c+\lambda},
\]
where \(k_c\) and \(n_c\) are the observed training successes and trials in cell \(c\), \(p_0(c)\) is the \(M1\) prediction for that cell, and \(\lambda\) is chosen by grouped cross-validation. For \(M2\), per-attribute contributions are defined interventionally by setting one \(\Delta_j\) to \(0\) while leaving the others unchanged and measuring the resulting change in predicted log-odds.

Table~\ref{tab:behavioural-robustness-metrics} uses the same held-out metric conventions as the main benchmark. Direct response choice agreement is agreement between the surrogate model's predicted choice and the model's observed choice. Score-based judge choice agreement is agreement between the score-based judge's reconstructed choice and the same surrogate-model choice target. Direct response attribute agreement is agreement between the attribute named by the model and the surrogate model's revealed driver, defined relative to the surrogate model's predicted choice on that row. Score-based judge attribute agreement is agreement between the score-based judge's largest-magnitude signed contribution and the same surrogate-model revealed driver.

Table~\ref{tab:behavioural-robustness-drivers} compares drivers across behavioural surrogates directly, using the same predicted-side convention. Each surrogate's driver is defined on the side predicted by that surrogate. If surrogate \(M\) predicts \(A\), its driver is the attribute with the largest contribution toward \(A\) and if it predicts \(B\), its driver is the attribute with the largest contribution toward \(B\).

\begin{table}[t]
\centering
\footnotesize
\small
\setlength{\tabcolsep}{4.8pt}
\renewcommand{\arraystretch}{1.08}
\begin{tabularx}{\columnwidth}{@{}>{\raggedright\arraybackslash}Xccc@{}}
\toprule
Metric & M0 & M1 & M2 \\
\midrule

Direct response choice agreement & 0.8039 & 0.8045 & 0.8056 \\
Score-based judge choice agreement & 0.7669 & 0.7641 & 0.7645 \\
Direct response attribute agreement & 0.6100 & 0.6094 & 0.6049 \\
Score-based judge attribute agreement & 0.6135 & 0.6174 & 0.6119 \\
\midrule
Held-out direct-choice NLL & 1.2068 & 1.1951 & 1.2035 \\
\bottomrule
\end{tabularx}
\caption{Headline robustness metrics under three behavioural surrogates.}
\label{tab:behavioural-robustness-metrics}
\end{table}

\begin{table*}[t]
\centering
\small
\setlength{\tabcolsep}{5pt}
\renewcommand{\arraystretch}{1.08}

\begin{tabular}{lccc}
\toprule
Scope & M0/M1 & M0/M2 & M1/M2 \\
\midrule
All 40 conditions & 0.9584 & 0.9250 & 0.9454 \\
Substantive themes only & 0.9583 & 0.9294 & 0.9448 \\
\bottomrule
\end{tabular}

\vspace{6pt}

\begin{tabular}{>{\raggedright\arraybackslash}p{2.7cm}cccccc}
\toprule
Theme & M0 NLL & M1 NLL & M2 NLL & \makecell{M0/M1\\driver agr.} & \makecell{M0/M2\\driver agr.} & \makecell{M1/M2\\driver agr.} \\
\midrule
Drugs & 1.3020 & 1.2846 & 1.3030 & 0.9344 & 0.8894 & 0.9063 \\
Policy & 1.0673 & 1.0677 & 1.0662 & 0.9769 & 0.9612 & 0.9719 \\
Software & 1.2511 & 1.2330 & 1.2412 & 0.9637 & 0.9375 & 0.9563 \\
Placebo packaging & 1.0761 & 1.0704 & 1.0743 & 0.9794 & 0.9300 & 0.9469 \\
Placebo label border & 1.0619 & 1.0562 & 1.0584 & 0.9375 & 0.9069 & 0.9456 \\
\bottomrule
\end{tabular}

\caption{Cross-surrogate agreement in recovered drivers and by-theme held-out fit. Driver-agreement numbers use the predicted-side surrogate convention described in the text.}
\label{tab:behavioural-robustness-drivers}
\end{table*}

Across themes, the headline metrics move only slightly under nearby surrogate choices. \(M1\) yields the cleanest robustness result: it modestly improves held-out fit relative to \(M0\) while leaving the choice-agreement and attribute-agreement conclusions essentially unchanged. \(M2\) remains supportive, but its driver assignments move more relative to \(M0\) and \(M1\), as reflected in lower cross-surrogate driver agreement, especially for \(M0/M2\). Taken together, these results indicate that the paper's main conclusion -- namely that held-out choices are substantially more recoverable than stated attributes -- does not depend on a single fragile surrogate specification.

\subsection{A structurally different six-attribute decision setting}
\label{app:hospital-setting}

\subsubsection{Benchmark description}

As a structurally different synthetic decision setting, we evaluated a \emph{hospital cyber-response} benchmark. The task objective was: ``Choose the better 12-month cyber-resilience plan for a mid-sized hospital with a small IT team.'' Each choice problem compared two plans on six ordered positive attributes: Threat reduction, Deployment speed, Staff sustainability, Service continuity, Audit readiness, and System portability. As in the main benchmark, we generated 400 training source families and 100 held-out test source families, rendered each family in four order/reversal variants, and sampled each rendering three times. We kept the same core direct-response and score-based judge protocol as in the main benchmark: the model reported both an \(A/B\) choice and a single most important attribute, while the judge returned one decisiveness score per visible attribute. For each model condition, we fit a separate behavioural model on the direct-response training choices and used it as the reference model on the held-out split. We evaluated GPT-5-mini NT and Qwen3-14B NT.

\begin{table*}[t]
\centering
\small
\renewcommand{\arraystretch}{1.08}
\begin{tabularx}{\textwidth}{@{}l*{5}{>{\centering\arraybackslash}X}@{}}
\toprule
Model & \makecell{Direct\\choice} & \makecell{Judge\\choice} & \makecell{Direct\\attribute} & \makecell{Judge\\attribute} & \makecell{Direct\\NLL} \\
\midrule
GPT-5-mini NT & \mci{0.847}{0.825}{0.866} & \mci{0.842}{0.820}{0.861} & \mci{0.644}{0.617}{0.671} & \mci{0.600}{0.572}{0.627} & \mci{0.958}{0.795}{1.126} \\
Qwen3-14B NT & \mci{0.890}{0.871}{0.906} & \mci{0.778}{0.753}{0.800} & \mci{0.560}{0.532}{0.588} & \mci{0.433}{0.405}{0.461} & \mci{0.704}{0.578}{0.842} \\
\bottomrule
\end{tabularx}
\caption{Headline held-out results for the hospital cyber-response setting. Direct response choice agreement is agreement between the behavioural model's predicted choice and the model's observed choice. Score-based judge choice agreement is agreement between the score-based judge's reconstructed choice and the same behavioural-model choice target. Direct response attribute agreement is agreement between the attribute named by the model and the behavioural model's revealed driver, defined relative to the model's observed choice. Score-based judge attribute agreement is agreement between the score-based judge's largest-magnitude signed contribution and the same behavioural-model revealed driver. Evaluation used 100 held-out source families, expanded to 400 rendered test prompts; with 3 samples per prompt, each agreement metric is based on 1{,}200 test responses per model. Held-out direct-choice NLL is computed over the 400 rendered test prompts.}
\label{tab:hospital-headline}
\end{table*}

\begin{table*}[t]
\centering
\footnotesize
\setlength{\tabcolsep}{4.8pt}
\renewcommand{\arraystretch}{1.08}
\begin{tabular*}{\textwidth}{@{\extracolsep{\fill}}lcccc@{}}
\toprule
\multicolumn{5}{@{}l}{\textit{Within-family pairwise agreement}}\\
Model & \makecell{Direct response\\choice} & \makecell{Direct response\\attribute} & \makecell{Score-based judge\\choice} & \makecell{Score-based judge\\attribute} \\
\midrule
GPT-5-mini NT & \mci{0.822}{0.778}{0.867} & \mci{0.616}{0.569}{0.663} & \mci{0.946}{0.919}{0.972} & \mci{0.677}{0.636}{0.722} \\
Qwen3-14B NT & \mci{0.861}{0.819}{0.900} & \mci{0.615}{0.571}{0.662} & \mci{0.867}{0.828}{0.906} & \mci{0.541}{0.497}{0.587} \\
\addlinespace[5pt]
\multicolumn{5}{@{}l}{\textit{Majority alignment with behavioural target}}\\
Model & \makecell{Direct response\\choice} & \makecell{Direct response\\attribute} & \makecell{Score-based judge\\choice} & \makecell{Score-based judge\\attribute} \\
\midrule
GPT-5-mini NT & \mci{0.890}{0.820}{0.950} & \mci{0.640}{0.540}{0.730} & \mci{0.860}{0.790}{0.920} & \mci{0.560}{0.460}{0.660} \\
Qwen3-14B NT & \mci{0.900}{0.840}{0.950} & \mci{0.510}{0.410}{0.610} & \mci{0.750}{0.660}{0.830} & \mci{0.340}{0.250}{0.430} \\
\bottomrule
\end{tabular*}
\caption{Perturbation summaries for the hospital setting, directly analogous to Figure~\ref{fig:reproducibility_alignment} in the main text. Each held-out source family contributes 12 realisations (4 prompt renderings \(\times\) 3 samples). Pairwise agreement is within-family pairwise agreement across these realisations. Majority alignment is the rate at which the dominant within-family response matches the relevant behavioural-model target, using the observed-choice-conditioned driver for attributes.}
\label{tab:hospital-perturbation}
\end{table*}

\subsubsection{Equalise-only occlusion summary}

As in the intervention analysis in the main text, \emph{equalise} sets one attribute to \emph{Medium} for both options while leaving the remaining attributes unchanged.

\begin{table*}[t]
\centering
\footnotesize
\renewcommand{\arraystretch}{1.08}
\begin{tabularx}{\textwidth}{@{}lccXX@{}}
\toprule
Model & \makecell{Choice-flip\\range} & \makecell{Stated-attribute\\flip range} & \makecell{Highest choice\\sensitivity} & \makecell{Highest stated-\\attribute sensitivity} \\
\midrule
GPT-5-mini NT & 0.093--0.238 & 0.169--0.430 & Threat reduction (0.238) & Threat reduction (0.430) \\
Qwen3-14B NT & 0.108--0.228 & 0.183--0.420 & Service continuity (0.228) & Threat reduction (0.420) \\
\bottomrule
\end{tabularx}
\caption{Equalise occlusion summary for the hospital setting. Each rate is computed over the 400 held-out rendered test prompts after equalising one attribute at a time.}
\label{tab:hospital-occlusion}
\end{table*}

Equalise occlusions therefore produced meaningful choice flips and even larger stated-attribute flips in both models. This again suggests that reported reasons are more labile than the underlying choice policy.

\subsubsection{Takeaway}

The same core dissociation survives in this structurally different six-attribute setting. In both models, held-out choices remained substantially recoverable from behaviour, while explicit attribute reports remained only partially aligned with the revealed driver. Because this setting has six possible attributes rather than four, the top-1 attribute-agreement task is also more demanding than in the base benchmark. The perturbation and equalise occlusion results are likewise supportive, extending the paper's evidence beyond the original shared four-attribute template.

\subsection{Additional targeted occlusion validations}
\label{app:additional-occlusion}

\subsubsection{Compact summary}

Table~\ref{tab:additional-occlusion-summary} summarises two additional four-attribute occlusion validations beyond the illustrative case shown in the main text. For compactness, we foreground the equalise intervention, which most closely matches the neutralisation logic used elsewhere in the paper. The table reports the baseline recovered order on the matched baseline rows, baseline-row choice accuracy, top-1 driver recovery, the largest equalise-only choice and stated-attribute flip rates, and the Spearman rank correlation between the baseline weight shares and the effect shares.

\begin{table*}[t]
\centering
\scriptsize
\setlength{\tabcolsep}{3.5pt}
\renewcommand{\arraystretch}{1.06}

\begin{tabularx}{\textwidth}{@{}
    >{\raggedright\arraybackslash}p{2.15cm}
    >{\raggedright\arraybackslash}X
    >{\centering\arraybackslash}m{1.25cm}
    >{\centering\arraybackslash}m{1.25cm}
    >{\centering\arraybackslash}m{1.55cm}
    >{\centering\arraybackslash}m{1.55cm}
    >{\centering\arraybackslash}m{1.05cm}
@{}}
\toprule
\multicolumn{1}{c}{Setting}
&
\multicolumn{1}{c}{Baseline order}
&
\makecell[c]{Base\\acc.}
&
\makecell[c]{Top-1\\rec.}
&
\makecell[c]{Top eq.\\choice flip}
&
\makecell[c]{Top eq.\\attribute flip}
&
\makecell[c]{Spearman\\$\rho$}
\\
\midrule
\makecell[l]{Qwen / policy\\(NT)}
&
Eff $>$ Comp $>$ Saf $>$ Impl
&
0.873
&
0.747
&
Eff (0.460)
&
Eff (0.513)
&
0.80
\\
\makecell[l]{Ministral / software\\(NT)}
&
Rel $>$ Cap $>$ Maint $>$ Adopt
&
0.811
&
0.538
&
Rel (0.380)
&
Rel (0.562)
&
1.00
\\
\bottomrule
\end{tabularx}

\caption{Two additional occlusion validations beyond the illustrative case in Figure~\ref{fig:occlusion_figure}. For compactness, the table foregrounds the equalise intervention, which most closely matches the neutralisation logic used elsewhere in the paper. Abbreviations are used, Comp = Compliance, Impl = Implementation Ease, Cap = Capability, Rel = Reliability, Maint = Maintainability, and Adopt = Adoption Ease. Spearman \(\rho\) is the rank correlation between baseline weight shares and equalise-effect shares.}
\label{tab:additional-occlusion-summary}
\end{table*}

\subsubsection{Synthesis}

The two additional settings show a useful contrast. In policy/Qwen3-14B NT, overall recovery is stronger: baseline-row choice accuracy is higher and top-1 driver recovery is substantially better. However, the relationship between learned baseline weights and intervention-effect shares is looser, with rank agreement of 0.80 rather than exact monotonic tracking. In software/Ministral-3-14B NT, overall recovery is weaker, but the learned ordering is tracked more cleanly by the occlusion effects, with exact rank agreement between the baseline weights and the effect shares. Taken together, these results show that the intervention evidence generalises across distinct behavioural profiles rather than depending on a single especially favourable setting.

\subsubsection{Baseline weight shares and occlusion-effect shares}

Table~\ref{tab:additional-occlusion-shares} reports the corresponding baseline weight shares and normalised occlusion effects for the two additional targeted validations.

\begin{table*}[t]
\centering
\scriptsize
\renewcommand{\arraystretch}{1.08}
\begin{tabularx}{\textwidth}{@{}>{\raggedright\arraybackslash}p{2.15cm}
>{\raggedright\arraybackslash}X
>{\raggedright\arraybackslash}X
>{\raggedright\arraybackslash}X
cc@{}}
\toprule
Setting & Baseline shares & Drop shares & Equalise shares & \makecell{\(\rho\)\\drop} & \makecell{\(\rho\)\\equalise} \\
\midrule
\makecell[l]{Qwen / policy\\(NT)}
& \makecell[l]{Eff: 0.399\\Comp: 0.223\\Saf: 0.216\\Impl: 0.162}
& \makecell[l]{Eff: 0.433\\Saf: 0.207\\Comp: 0.183\\Impl: 0.176}
& \makecell[l]{Eff: 0.446\\Saf: 0.201\\Comp: 0.192\\Impl: 0.161}
& 0.80
& 0.80 \\
\makecell[l]{Ministral / software\\(NT)}
& \makecell[l]{Rel: 0.384\\Cap: 0.282\\Maint: 0.195\\Adopt: 0.139}
& \makecell[l]{Rel: 0.389\\Cap: 0.309\\Maint: 0.183\\Adopt: 0.119}
& \makecell[l]{Rel: 0.373\\Cap: 0.304\\Maint: 0.189\\Adopt: 0.134}
& 1.00
& 1.00 \\
\bottomrule
\end{tabularx}
\caption{Baseline weight shares and normalised occlusion-effect shares for the two additional targeted validations. Theme-specific abbreviations follow Table~\ref{tab:theme-slots}: Eff = Effectiveness, Comp = Compliance, Impl = Implementation Ease, Cap = Capability, Rel = Reliability, Maint = Maintainability, and Adopt = Adoption Ease. The Qwen/policy setting shows a small reshuffling below the top attribute, whereas the Ministral/software setting tracks the learned ordering almost perfectly despite weaker overall recovery.}
\label{tab:additional-occlusion-shares}
\end{table*}

\subsection{Statistical and implementation details}
\label{app:stats-impl}

\subsubsection{Confidence intervals and bootstrap units}

For the appendix tables introduced here, 95\% confidence intervals for agreement rates are Wilson score intervals using the relevant denominator. Held-out direct-choice NLL intervals are computed by nonparametric bootstrap over held-out rendered test prompts, resampling rendered prompts with replacement and recomputing the mean binomial NLL on each resample. Perturbation summaries use a family bootstrap over held-out source families: source families are resampled with replacement, and the within family pairwise-agreement and majority alignment summaries are recomputed on each bootstrap replicate.

\subsubsection{Behavioural robustness fitting details}

Each behavioural surrogate is fit separately for each theme \(\times\) model family \(\times\) setting, using only direct-response training choices from the corresponding training split (score-based judge outputs are never used for fitting). All three surrogates are fit on rendered-row binomial counts, aggregating the three direct response samples for each rendered prompt into successes and trials. \(M0\) is the baseline unpenalised additive fit used in the main paper. \(M1\) is fit on the same training data and defaults to an unpenalised fit; when the unpenalised fit is numerically unstable, a ridge penalty is selected by grouped five-fold cross-validation with source family as the grouping unit, choosing the smallest \(\lambda\) attaining the best mean held-out NLL over the grid \(\{0,10^{-6},10^{-4},10^{-2},10^{-1}\}\). \(M2\) is a cell-lookup model over the full attribute-difference cell, shrunk toward an \(M1\) prior. Its shrinkage parameter is selected by the same grouped five-fold family-level cross-validation over \(\lambda \in \{0.5,1,2,5,10,20,50,100\}\), again taking the smallest \(\lambda\) with the best mean held-out NLL.

\subsubsection{Tied-attribute judge-score diagnostic}

As a diagnostic of the raw score-based judge outputs, we examined attributes on which the two options have the same level. In GPT-5-nano T, GPT-5-mini T, and Qwen3-14B T, tied attributes are almost always assigned zero raw score, with non-zero rates between \(0.12\%\) and \(0.42\%\). In GPT-5-mini NT, Ministral-3-14B T, and Qwen3-14B NT, tied attributes often receive positive raw scores, with non-zero rates between \(87.49\%\) and \(88.62\%\), although their mean scores remain lower than those of non-tied attributes (approximately \(0.3\) versus \(0.6\)). Thus, in some settings the raw scores mix general attribute salience with case-specific decisiveness. This does not affect the reported judge-choice or judge-driver metrics, because tied attributes receive sign zero before aggregation.

\subsection{Prompts}
\label{app:prompts}
\subsubsection{Direct response prompt}

\begin{verbatim}
You are optimizing [decision objective].
Choose exactly one option and then give the single most important attribute for that
choice.
Respond in exactly one line using this format:
<A or B>. <attribute>.
Attributes: [attribute list].
Do not give any explanation or additional commentary.
Options:

[Option A profile]

[Option B profile]
\end{verbatim}

\subsubsection{Score-based judge prompt}

\begin{verbatim}
You are an evaluator. Score how decisive each attribute difference is for
[decision objective].
Return a number between 0 and 1, where 0 = no impact and 1 = fully decisive.
Attributes: [attribute list].
If an attribute is not shown, set its score to 0.

[Option A profile]

[Option B profile]

Return exactly one line per attribute and nothing else.
Each line must be:
<attribute>=<score>

Use one single numeric score per attribute.
Do not use brackets, intervals, commas, or explanations.
Follow this format, replacing <score> with a number between 0 and 1:

[attribute 1]=<score>
[attribute 2]=<score>
...
[attribute N]=<score>
\end{verbatim}

\subsection{Reproducibility}

\paragraph{Full model names and hyperparameters}

\begin{itemize}
\item \textbf{GPT-5-mini (NT/T)} gpt-5-mini: OpenAI default sampling settings, reasoning\_effort=minimal/low, max\_tokens=4000.
\item \textbf{GPT-5-nano (NT/T)} gpt-5-nano: OpenAI default sampling settings, reasoning\_effort=minimal/low, max\_tokens=4000.
\item \textbf{Qwen3-14B (NT/T)} Qwen/Qwen3-14B: temperature=1.0, top\_p=0.95, max\_tokens=2048.
\item \textbf{Ministral-3-14B (NT)} mistralai/Ministral-3-14B-Instruct-2512: temperature=1.0, top\_p=0.95, max\_tokens=2048.
\item \textbf{Ministral-3-14B (T)} mistralai/Ministral-3-14B-Reasoning-2512: temperature=1.0, top\_p=0.95, max\_tokens=2048.
\end{itemize}
\end{document}